\documentclass[conference]{IEEEtran}

\usepackage{amsmath}
\usepackage{balance}
\usepackage[noadjust]{cite}
\usepackage{gensymb}
\usepackage{threeparttable}
\usepackage{soul}
\usepackage{caption}
\usepackage{subcaption}

\ifCLASSINFOpdf
  \usepackage[pdftex]{graphicx}
\else
\fi
\usepackage{array}

\usepackage{multirow}
\usepackage{xcolor}
\begin{document}
\title{Realtime Facial Expression Recognition: Neuromorphic Hardware vs. Edge AI Accelerators\\
}
\author{\IEEEauthorblockN{Heath Smith, James Seekings, Mohammadreza Mohammadi, Ramtin Zand}
\IEEEauthorblockA{Department of Computer Science and Engineering, University of South Carolina, Columbia, SC 29208, USA\\
e-mail: smithmh6@email.sc.edu, seekingj@email.sc.edu, mohammm@email.sc.edu, ramtin@cse.sc.edu
}
}

% \author{\IEEEauthorblockN{Heath Smith}
% \IEEEauthorblockA{\textit{Computer Science and Engineering} \\
% \textit{University of South Carolina}\\
% Columbia, USA \\
% smithmh6@email.sc.edu}
% \and
% \and
% \and
% \IEEEauthorblockN{Ramtin Zand}
% \IEEEauthorblockA{\textit{Computer Science and Engineering} \\
% \textit{University of South Carolina}\\
% Columbia, USA \\
% ramtin@cse.sc.edu}
% }

\maketitle


\maketitle

\begin{abstract}
The paper focuses on real-time facial expression recognition (FER) systems as an important component in various real-world applications such as social robotics. We investigate two hardware options for the deployment of FER machine learning (ML) models at the edge: neuromorphic hardware versus edge AI accelerators. Our study includes exhaustive experiments providing comparative analyses between the Intel Loihi neuromorphic processor and four distinct edge platforms: Raspberry Pi-4, Intel Neural Compute Stick (NSC), Jetson Nano, and Coral TPU. The results obtained show that Loihi can achieve approximately two orders of magnitude reduction in power dissipation and one order of magnitude energy savings compared to Coral TPU which happens to be the least power-intensive and energy-consuming edge AI accelerator. These reductions in power and energy are achieved while the neuromorphic solution maintains a comparable level of accuracy with the edge accelerators, all within the real-time latency requirements.

\end{abstract}

\begin{IEEEkeywords}
Facial Expression Recognition, Neuromorphic Computing, Spiking Neural Networks (SNNs), Convolutional Neural Networks (CNNs), Edge Computing.
\end{IEEEkeywords}

% \fi
\IEEEpeerreviewmaketitle

\section{Introduction}

Facial expression recognition is a popular and challenging area of research in machine learning (ML) applications \cite{FER_Survey}. Facial expressions are critical to human communication and allow us to convey complex thoughts and emotions beyond spoken language. Using computer vision and ML systems for facial expression classification can be beneficial to many real-world applications. The complexity of facial expressions, however, creates a difficult problem for computer vision systems. Moreover, the applications that require facial expression, such as social robotics \cite{social_robots} and sign language translation \cite{mccullough2005neural}, typically demand real-time operation at the edge which can cause important challenges due to the resource and energy constraints of edge platforms \cite{edge_survey}.

The surge in demand for specialized hardware for artificial intelligence (AI) applications has resulted in a rapidly expanding industry for edge AI accelerators. Anticipating this trend, several companies have developed their own specialized accelerators. The NVIDIA Jetson Nano is a low-cost development board for ML applications that is equipped with both GPU and CPU cores. The Intel Movidius Neural Compute Stick 2 (NCS2) is a small, low-power USB co-processor that enables ML acceleration by the Myriad vision processing unit (VPU). Google's Coral Edge TPU is another device that leverages tensor processing units (TPUs) to accelerate ML applications. The coral TPU is used as a co-processor on Coral's Dev Board, as well as a USB accelerator that can be integrated with tiny computers such as Raspberry Pi. Despite these advancements, the development of efficient ML models tailored for resource and energy-constrained edge devices remains a challenge that is exacerbated by the increasing complexity and size of network architectures \cite{edge_survey}. To address these challenges, there has been a specific focus on leveraging automated ML (AutoML) \cite{hutter2019automated,baymurzina2022review} techniques to develop ML models that are tailored and optimized for deployment on edge devices. While recent research in this field \cite{FER_GLSVLSI,gupta2020accelerator,reidy2023efficient} has shown promising outcomes, there is still room for further exploration and enhancement of edge-centric ML models.

On the other hand, neuromorphic systems representing the third generation of ML systems \cite{lobo2020spiking} have attracted considerable attention in recent years and exhibited intriguing outcomes for various edge computing applications \cite{schuman2022opportunities, mohammadi2022static,9395703}. Neuromorphic computing systems include properties such as sparse and low-precision computation as well as event-based and asynchronous operation that make them intrinsically suitable for low-energy ML tasks \cite{9395703, LessonsLoihi}. 

In this paper, we first employ AutoML techniques to develop convolutional neural network (CNN) models that are specifically optimized for being executed on four distinct edge platforms: Raspberry Pi-4, Intel NCS2, NVIDIA Jetson Nano, and Coral TPU. Subsequently, we convert these CNN models to equivalent spiking neural networks (SNNs) for deployment on an Intel Loihi neuromorphic processor \cite{intelloihi}. Finally, we provide comprehensive experimental results and comparative analyses between the edge AI accelerators and neuromorphic hardware in terms of accuracy, latency, power, and energy consumption. The results unveil important trade-offs and nuances that can be crucial for the practical deployment of facial expression recognition systems for real-world applications.

\section{DNN and SNN Model Development}
\label{sec:DNN-SNN}

\subsection{Develop and Optimize CNN Models}
%Here, we leverage an edge-centric approach for network architecture search \cite{FER_GLSVLSI}, which creates CNN models that are optimized for deployment on edge AI accelerators. The approach aims to achieve a balance between accuracy, latency, and power dissipation through an optimization mechanism that hierarchically adds new metrics to the cost function to narrow down the search space starting with optimizing the \textit{accuracy} first and ending up with searching the design space to find the model with best \textit{accuracy/PDP} value, where PDP is power-delay-product.   

Here, we employ an edge-centric methodology for conducting network architecture search, which is specifically designed to develop CNN models that are highly efficient when deployed on edge AI accelerators. The primary objective of this approach is to achieve a balance between accuracy, latency, and power consumption. This is accomplished through a systematic optimization process that progressively integrates new metrics into the cost function. The optimization begins with prioritizing \textit{accuracy}, subsequently progressing to include \textit{latency}, and ultimately ending the exploration of the design space to identify the model with the optimal \textit{accuracy/PDP} ratio, where PDP represents power-delay-product. Here, we use Hyperopt \cite{bergstra2015hyperopt} library to perform automated hyperparameter tuning based on the aforementioned cost functions using Bayesian optimization techniques.

\begin{table}
\caption{Network configuration settings.}
\label{table:Netconf}
\resizebox{\columnwidth}{!}{%
\begin{tabular}{clclclc}
\hline
Parameters &  & Description &  & Range &  & Step \\ \hline 
Block &  & \# of blocks in the network &  & 2-4 &  & 1 \\
K1 &  & \# of kernels in the 1st block &  & 6-16 &  & 2 \\
K2 &  & \# of kernels in the 2nd block &  & 24-32 &  & 4 \\
K3 &  & \# \# of kernels in the 3rd block &  & 36-48 &  & 4 \\
K4 &  & \# \# of kernels in the 4th block &  & 52-64 &  & 4 \\
FC1 &  & \# of units in the 1st FC layer &  & 100-120 &  & 5 \\
FC2 &  & \# of units in the 2nd FC layer &  & 80-100 &  & 5 \\ \hline
\end{tabular}}
\label{table:net_info}
\end{table}

The structure of the CNN models used in this paper is based on the VGG architecture \cite{vgg}. Each model is composed of multiple VGG blocks, each of which includes two convolution layers with 3$\times$3 kernel dimensions and a stride of 1. The parameters that can be adjusted to fine-tune the CNN architecture are outlined in Table \ref{table:Netconf}, and described as below: 
\begin{itemize}
    \item The number of VGG blocks ($Block$) can range between 2 and 4, allowing for varying network depths.
    \item The number of kernels within each VGG block can be altered.
    \item The number of nodes in each fully connected (FC) layer is variable.
\end{itemize}

\subsection{CNN to SNN Conversion}
To deploy a pre-trained CNN model on neuromorphic hardware, i.e., Loihi \cite{intelloihi} in this paper, it must be first converted to a spiking model. When evaluating a typical convolutional network for image recognition tasks, data enters the network as static frames. Static images must be converted to spike trains before being processed by SNNs \cite{StaticEncode}. Therefore, we defined an additional convolutional layer at the beginning of the network model which has a kernel size of one and is executed off-chip. By running this initial layer outside of the Loihi chip, the static input images can be converted into spikes, which are then fed into the main network architecture that is on the Loihi chip. The off-chip input layer utilizes a rate-based encoding method to convert the static input images into output spiking signals. There is also an additional training step required when preparing models for Loihi. Models trained with standard ReLU activation functions cannot be directly deployed on Loihi, so we must first replace all of the ReLU activation functions with a software abstraction of Loihi's spiking activation function, which models a Leaky-Integrate-and-Fire (LIF) neuron. Once the activation functions are replaced, a second training step is conducted to allow the spiking neurons to learn the characteristics of the dataset.

%Once we have an off-chip layer to convert the image data into spikes, we can begin refactoring the rest of the network to fit on the Loihi chip. 

In addition, because max pooling cannot be directly implemented in SNNs, alternative mechanisms must be considered \cite{snntoolbox}. While it is possible to implement pooling off-chip, it can lead to significant data transfer costs to move data to and from the Loihi. Instead of max or average pooling, we can remove pooling layers completely and replace it with a strided convolution layers. By doing so, we can keep the entire network on the Loihi chip. The strided convolution can also offer some advantages; because pooling is a fixed mathematical operation, there is no learning involved in a pooling layer. By using strided convolution, we allow the network to learn how to pool the data \cite{ayachi2020strided}. After refactoring the convolutional portion of the network, dense layers can be directly converted into spiking layers.

%\hl{In our designed architecture, we have two fully connected layers after a flattening layer.} 

%While fully connected layers are able to run on the Loihi chip, we must take our last output layer and run it off-chip. This allows Loihi to send the output spikes to the classification layer and the classification predictions can be calculated off-chip which reduces the overall number of off-chip calculations required. 

%The full architecture of the converted spiking network can be observed in Figure \ref{fig:loihi_architecture}. We can see that the dimensionality of the input is gradually changed as the input data moves through the network. We can also observe the absence of the pooling layers in favor of strided convolution, as well as the off-chip input and output layers.

After converting CNNs to a spiking network, we must take memory limitations into consideration. Every Loihi core has the capability to accommodate up to 1,024 neurons, and a single chip comprises 128 such cores \cite{intelloihi}. Because of these limitations, we must break apart the convolutional layers into separate blocks that can operate in parallel across multiple cores of the chip. Loihi is able to support partially filled cores, so we do not need to have a perfect fit on each layer, and making sure that there are not more than $1024$ neurons assigned to a single core is sufficient. The block size of the layers can be set by a 3-element tuple, in the form of rows$\times$columns$\times$channels. The product of the number of rows, columns, and channels must not exceed $1024$ for a single block. Finally, prior to running the model on Loihi, we need to insert hardware-level probes to measure latency, power, and energy consumption during inference. Intel provides access to these probes through its neuromorphic developer kit. The probes should be activated before building the network on the Loihi chip.

%In this experiment, we would like to collect data on energy consumption and latency, so we will be using the energy probe and execution time probes provided by Intel. 
%These probes must be activated before building the network on the Loihi chip. %In these tests, we probed the energy and power consumption of the entire network for the full duration of running inference on the Loihi chip. 

%After the model is converted and we have the probes set up, the model can be compiled and executed on Loihi. Due to the time cost of transferring data to and from the Loihi chip, five images were chosen to evaluate the model performance on Loihi. Each of the five images was exposed to the network for a period of $50$ milliseconds and this process was repeated for both grayscale and edge-detected input images. 

\section{Experimental Setup}

\subsection{Dataset}
We use the \textit{Extended Cohn-Kanade} (CK+) \cite{ckplus} dataset including seven facial expressions: \textit{happiness}, \textit{sadness}, \textit{surprise}, \textit{anger}, \textit{contempt}, \textit{disgust}, or \textit{fear}. %Figure \ref{fig:ck_samples} shows some sample images from the dataset that are taken in a controlled environment. 
In this paper, we resized the images to $48\times48$ pixels and converted them to grayscale before training. By reducing the size and the number of color channels in the images, we can lower the number of trainable parameters needed in the CNN models with minimal or no loss of accuracy. 

% \begin{figure}[!b]
%     \centering
%     \includegraphics[width=3in]{Figures/ck_sample.png}
%     \caption{Sample images taken from the CK+ dataset \cite{ckplus}.}
%     %\vspace{-11mm}
%     \label{fig:ck_samples}
% \end{figure}

\subsection{Edge AI Accelerators}

\begin{figure}
    \centering
    \includegraphics[width=3.3in]{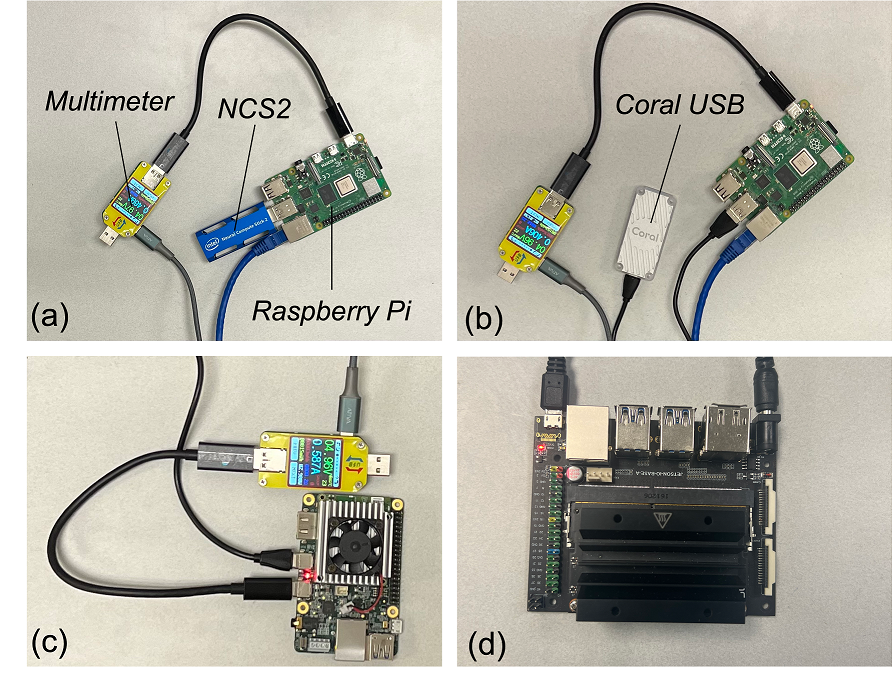}
    \caption{Experimental setup. (a) Pi + NCS2 (b) Pi + Coral TPU (c) Coral Dev board (d) Jetson Nano.}
    \label{fig:expri_hw}
\end{figure}

Herein, we provide a comprehensive comparison between Loihi and various edge AI accelerators for the facial expression recognition task in terms of accuracy, latency, power efficiency, and energy consumption. In particular, we employ four devices including Raspberry Pi, Google Coral TPU (comprising both the Development Board and USB variants), Intel Movidius Neural Compute Stick 2 (NCS2), and Nvidia Jetson Nano. These setups are illustrated in Figure \ref{fig:expri_hw}.

%Here, we compare Loihi against various edge AI accelerators for facial expression recognition application in terms of accuracy, latency, power, and energy consumption. In particular, we use four devices including Raspberry Pi, Google Coral TPU (both Development board and USB), Intel Movidius neural compute stick 2 (NCS2), and Nvidia Jetson Nano, as shown in Fig. \ref{fig:expri_hw}. %Each edge device has its own characteristics and deployment procedures as described in the following.

%to assess how they perform in terms of accuracy, latency, and power consumption. Different parameters are necessary to run the models on each edge device. We take advantage of each hardware's lowest precision that is feasible. For all models, we set the batch size to one. 
%We utilize TFlite models with FP32 precision for the Raspberry Pi4,  TFLite models with INT8 precision for both Coral products,  We use OpenVino models with FP16 precision for the NCS2. For our experiments, we employed OpenVINO 2022.2. TensorRT models with FP16 precision are used in the development of the Jetson Nano. We export the models from TensorFlow to Onnx and then import them into TensorRT using OnnxParser. For the Jetson Nano, we use Ubuntu 18.04. For the sake of completeness, we evaluate the models on Jetson using Max-N and low power modes.

\begin{figure*}
    \centering
    \includegraphics[width=5.8in]{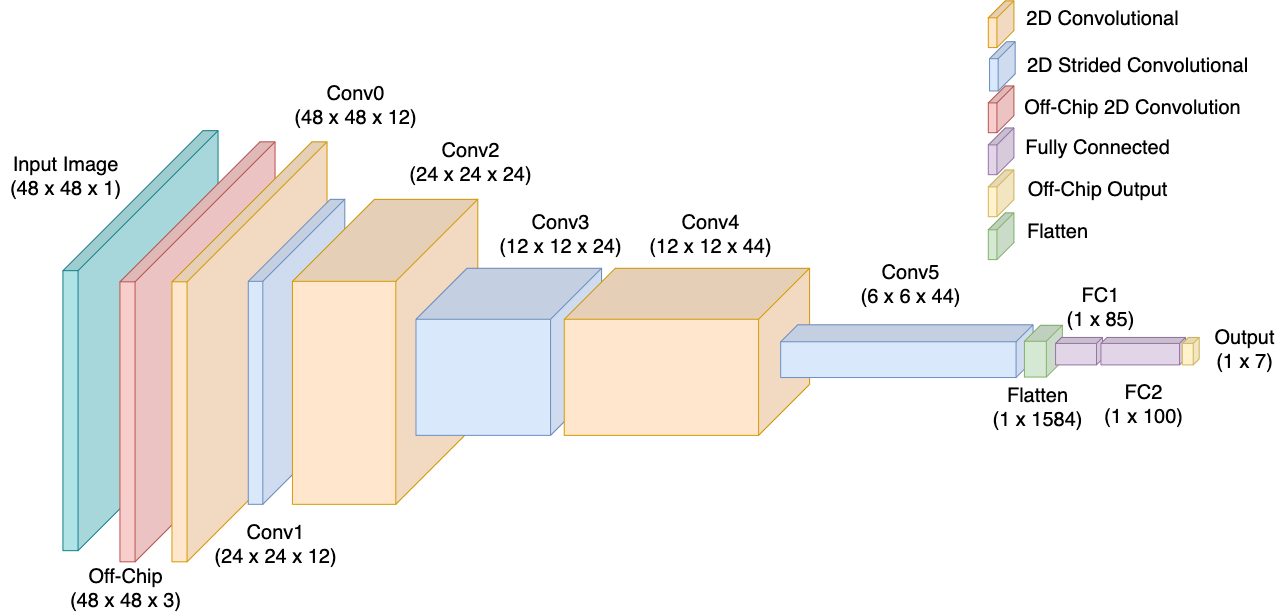}
    %\vspace{-2mm}
    \caption{Spiking network architecture as deployed on Loihi chip.}
    %\vspace{-11mm}
    \label{fig:loihi_architecture}
\end{figure*}

\subsubsection{Nvidia Jetson Nano}

%\textcolor{red}{The NVIDIA Jetson Nano 
%is a low-cost development board for ML applications. The Jetson Nano is a modular computer with a Tegra X1 SoC that combines an ARM A57 quad-core, a 1.43 GHz CPU, and four distinct 32-CUDA core processing blocks (128 CUDA cores total) within a Maxwell architecture. It also comes with 4 GB of RAM. %memory. Nvidia provides two operating modes for Jetson Nano, i.e., 5W and Max-N, that can be configured through a software interface. In the 5W mode, also known as low power mode, only two cores of the ARM A57 are powered on, the clock frequency is restricted to 0.9 GHz and the GPU's clock frequency is limited to 0.64 GHz. However, in the Max-N mode, the Arm 57's four active cores operate at a clock frequency of 1.5 GHz, while the GPU's clock frequency is 0.92 GHz. TensorRT models with 16-bit floating point (FP16) precision are used for Jetson Nano. We export the models from TensorFlow to ONNX and then import them into TensorRT using OnnxParser.For the purpose of optimizing and quantizing deep learning models, Jetson Nano employs NVIDIA TensorRT as its main driver.}

The Jetson Nano is a development board with a Tegra X1 System-on-Chip (SoC) that integrates an ARM A57 quad-core CPU with four 32-CUDA core processing units. It also includes 4 GB of memory. Jetson Nano can operate in two modes: (1) low power mode (Jetson-L), in which only two CPU cores of the ARM A57 are powered on at the clock frequency of 0.9 GHz, and the GPU's clock frequency is limited to 0.64 GHz; (2) high power mode (Jetson-H), in which all the four CPU cores are activated with a clock frequency of 1.5 GHz, while the GPU's clock frequency is set to 0.92 GHz. To deploy CNN models on the Jetson Nano platform, the workflow involves exporting these models from TensorFlow to ONNX format, followed by importing them into TensorRT using the OnnxParser tool. %For the purpose of optimizing and quantizing deep learning models, Jetson Nano employs NVIDIA TensorRT as its main driver.

% \vspace{-1cm}
\subsubsection{Intel Neural Compute Stick 2 (NCS2)}

The NCS2 leverages the Intel Movidius X Vision Processing Unit (VPU), which includes 16 programmable cores along with a neural compute engine. NCS2 is also equipped with 4GB of memory with a maximum frequency of 1.6 GHz. To facilitate the deployment of ML models on the NCS2, Intel provides a library called OpenVINO that includes a model optimizer to transform the models to a format supported by NCS2.

%Following model optimization for \textcolor{violet}{the }Intel NCS2, we perform inference using OpenVINO's built-in inference engine API to measure the accuracy and latency.

\subsubsection{Google Coral TPU}
%he Google edge TPU \cite{coral_overview}is a different accelerator built for applications \textcolor{violet}{that use }%using machine learning. 
Both the Coral Development Board and USB accelerator leverage tensor processing unit (TPU) as a co-processor within a system-on-module (SoM) architecture to accelerate ML workloads. To deploy ML models on Coral Edge TPU, all the models must be converted to the TensorFlow Lite format and quantized to 8-bit integer values.

\begin{figure*}
    \centering
    \begin{subfigure}{\textwidth}
        \centering
        \includegraphics[width=6in]{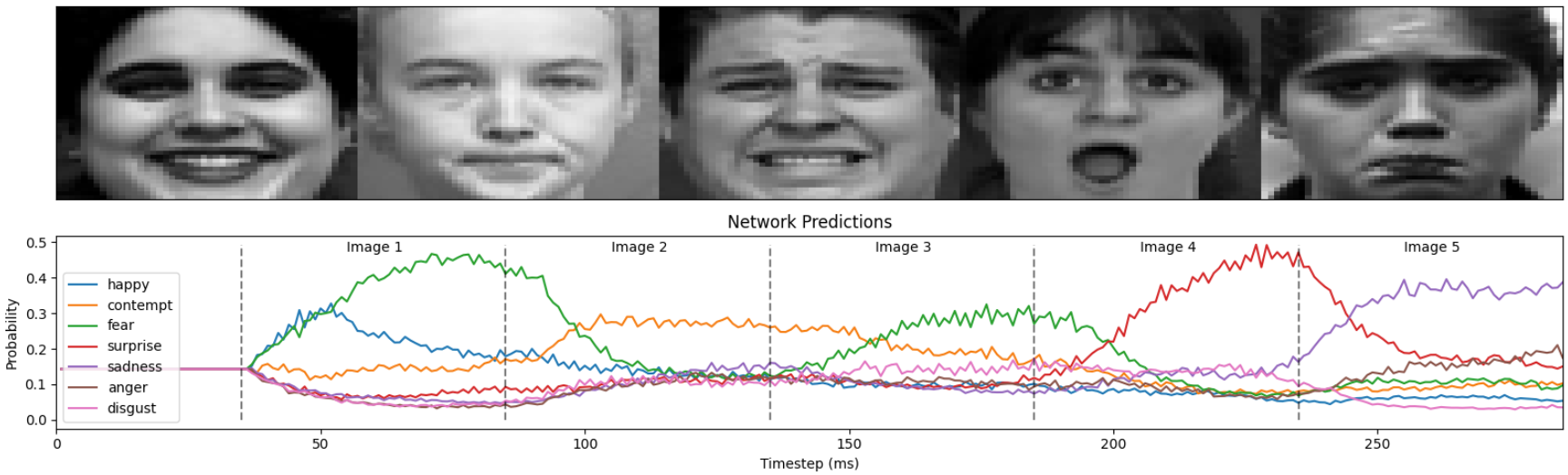}
        \caption{}
    \end{subfigure}
    \begin{subfigure}{0.9\textwidth}
       \centering
       \includegraphics[width=6in]{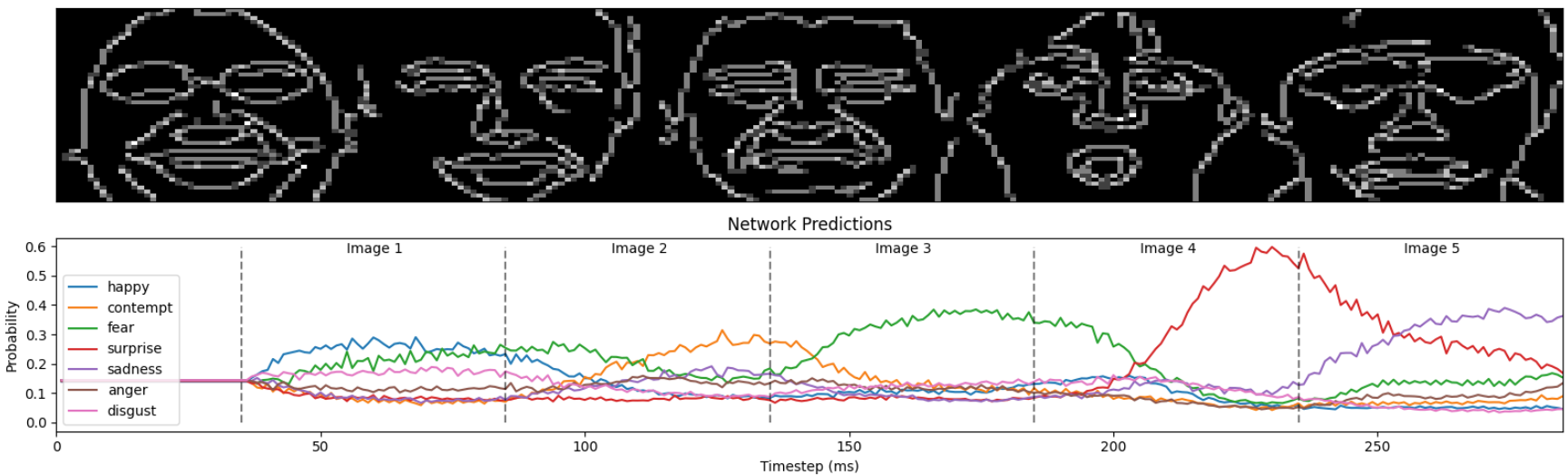} 
       \caption{}
    \end{subfigure}
    %\vspace{-2mm}
    \caption{Inference results from five images tested on Loihi: (a) grayscale image, and (b) edge-detected images.}
    %\vspace{-11mm}
    \label{fig:loihi_inference}
\end{figure*}

\subsection{Neuromorphic Hardware: Loihi}
Herein, we use the Intel Loihi chip \cite{intelloihi} in our experiments and compare its performance against the aforementioned edge AI accelerators. Loihi is a neuromorphic chip that provides a variety of features including hierarchical connectivity, dendritic compartments, synaptic delays, and programmable synaptic learning rules. Each Loihi chip consists of a many-core mesh, which is comprised of 128 neuromorphic cores combined with three embedded processors. The chip also includes hardware interfaces to allow communication with other Loihi chips. Each of the 128 cores in the Loihi chip contains 
1,024 spiking neural units which are grouped into trees that make up a neuron. The neurons found in the Loihi chip imitate biological neurons, which are well-suited for processing time-dependent input signals. %Convolution models can be deployed on Loihi in the form of spiking CNNs. 
The Loihi chip also offers hardware abstractions in Intel's neuromorphic API which allows users to measure energy and power usage while running inference on the chip.

\section{Results and Discussions}

% \begingroup
%     \setlength{\tabcolsep}{6pt} % Default value: 6pt
%     \renewcommand{\arraystretch}{1.5} % Default value: 1
%     \begin{table}
%         \centering
%                 \caption{Accuracy of the CNN and converted SNN deployed on the Loihi chip.}
%         \label{tab:loihi_performance}
%         \begin{tabular}{lccc}
%             \textbf{Model}  & \textbf{Grayscale} & \textbf{Edge Detection}\\
%             \cline{1-3}
%             CNN      & $99.49\%$ & $84.72\%$ \\
%             SNN on Loihi & $0.0\%$ & $94.25\%$ \\
%             \hline
%         \end{tabular}

%     \end{table}
% \endgroup

% \begin{figure*}
%     \centering
%     \includegraphics[width=6in]{Figures/loihi_result_gray.png}
%     %\vspace{-2mm}
%     \caption{Inference results from 5 grayscale images tested on Loihi.}
%     %\vspace{-11mm}
%     \label{fig:loihi_gray_inference}
% \end{figure*}

In the first stage of our experiments, we used the optimization approach discussed in Section II.A to find the best CNN model architecture for each edge device that achieves a balance between accuracy, latency, and power consumption. Table \ref{table:acc_pdp_all_devices} lists the hyperparameters for the best models selected by the optimization algorithm. As it can be observed, none of the models include the fourth VGG black, which can be interpreted as the accuracy gain of making the model deeper is not worth the increased latency and power dissipation. The model selected for deployment on Loihi is the deepest model with three VGG blocks compared to the two VGG blocks for the models deployed on edge AI accelerators. The need for using a deeper network for neuromorphic hardware can be associated with the potential loss of accuracy that can occur when pre-trained CNNs are converted to SNN as also investigated in \cite{snntoolbox, IGSCdeploy}. The full architecture of the converted spiking network can be observed in Figure \ref{fig:loihi_architecture}. We can see that the dimensionality of the input is gradually changed as the input data moves through the network. We can also observe the absence of the pooling layers in favor of strided convolutions.%, as well as the off-chip input and output layers.

%After the CNN to SNN conversion is complete, we have compiled and executed the SNN model on Loihi. %Due to the time cost of transferring data to and from the Loihi chip, five images were chosen to evaluate the model performance on Loihi. Each of the five images was exposed to the network for a period of $50$ milliseconds and this process was repeated for both grayscale and edge-detected input images. 

\subsection{Accuracy}

In Figure \ref{fig:loihi_inference}, a visual representation of the SNN network is shown as it processes five images from the CK+ dataset, each of which is exposed to the network for 50 milliseconds. The figure includes the traces that represent the probability of each of the seven facial expression classification classes in the CK+ dataset. We can observe a propagation delay from the time an input is first presented to the network until the time when the probability is returned. This is due to the amount of time it takes for spikes to propagate through the network. To explore the effect of input sparsity on the network, which has been shown to increase performance in SNNs \cite{IGSCdeploy}, we performed edge detection on the input images which can be seen in Fig \ref{fig:loihi_inference} (b). %\textcolor{blue}{\st{It can also be observed that the output of the network is highly unstable over time. The output does not show any consistency as it fluctuates from high to low values rapidly, making it difficult to achieve a correct prediction. This could be caused by the lack of sparsity in the input images, which can negatively affect the performance of the SNNs, as also investigated in the literature. Therefore, to enhance the sparsity in the input images, we have added performed an edge detection on the input images and applied them to the network. Figure 5 shows the classification results on the edge detected images.}} 
There appears to be less noise in the probability signals for the edge-detected images as compared to the signals in the gray-scale images, % in Figure \ref{fig:loihi_gray_inference}, \textcolor{blue}{\st{which has led to more accurate classification results.}} 
which can be seen by the smoother probability traces. This can lead to a more accurate classification. For instance, as shown in the figure, the first image of the grayscale is incorrectly classified as \textit{fear}, whereas with the edge-detected images, it is correctly classified as \textit{happy}, despite being closely tied with \textit{fear}. The entire test dataset was evaluated using the \textit{Loihi Spiking Rectified Linear Unit} activation function provided by the NengoDL conversion tool \cite{bekolay2014nengo}. The model was able to achieve an accuracy of 94.79\% and 97.40\% for grayscale and edge-detected images, respectively.

\begin{figure}
    \centering
    \includegraphics[width=3.4in]{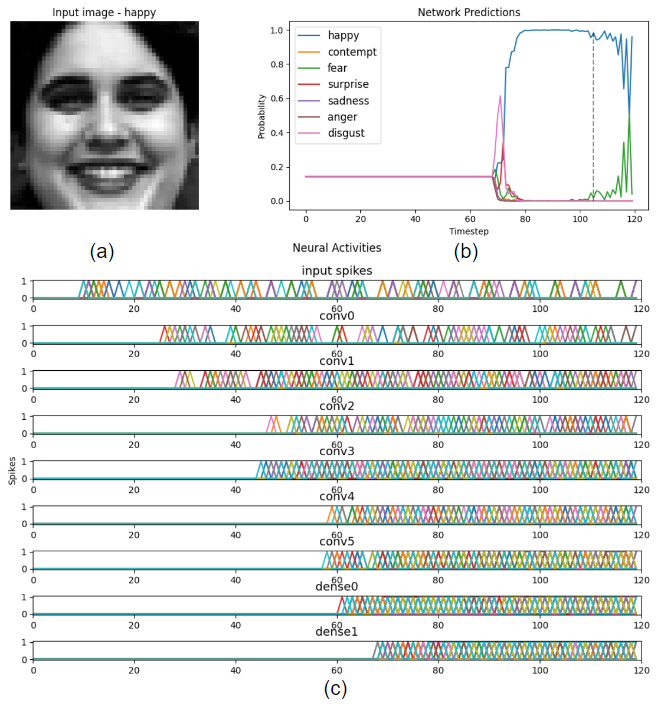}
    \vspace{-2mm}
    \caption{(a) Sample grayscale image, (b) corresponding network predictions, and (c) neural activity in different layers.}
    %\vspace{-11mm}
    \label{fig:loihi_gray_neurons}
\end{figure}

\begin{figure}
    \centering
    \includegraphics[width=3.4in]{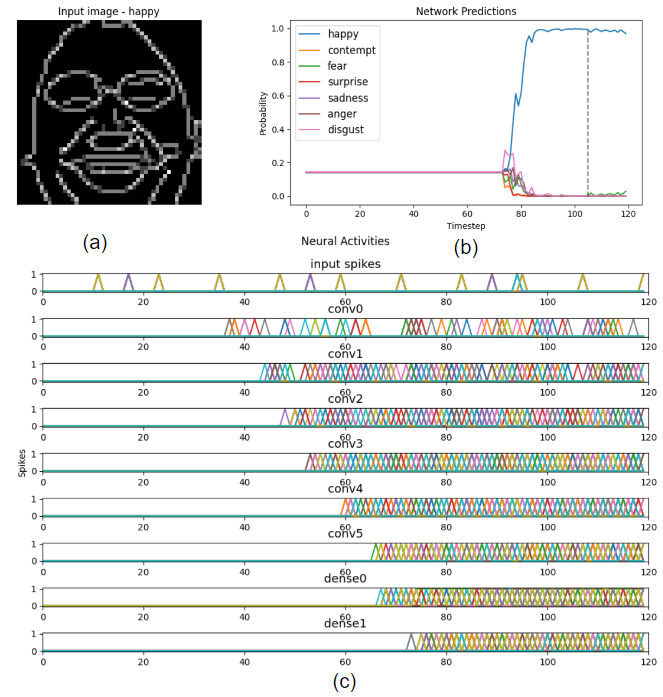}
    \vspace{-2mm}
    \caption{(a) Sample edge-detected image, (b) corresponding network predictions, and (c) neural activity in different layers.}
    %\vspace{-11mm}
    \label{fig:loihi_edge_neurons}
\end{figure}

To investigate the reason for the accuracy gain in the edge-detected images, we show the neural activities of different layers of the SNN model for a representative gray-scale image and its corresponding edge-detected version in Fig. \ref{fig:loihi_gray_neurons} and \ref{fig:loihi_edge_neurons}, respectively. In NengoDL, spike probes were placed in each layer of the SNN, and are shown with the first convolutional layer at the top, and the last fully connected layer at the bottom of Fig. \ref{fig:loihi_gray_neurons} (c) and Fig. \ref{fig:loihi_edge_neurons} (c). The probes were set to monitor a sample size of 50 neurons in each layer. From the figures, we can observe that due to the lack of sparsity in the input data, the network has difficulty determining what features are important and thus the overall activity level is much higher, leading to noisier outputs shown in Fig. \ref{fig:loihi_gray_neurons} (b). When neurons fire more sparsely, it is easier for the network to discern important features, which can lead to more robust classification as shown in Fig. \ref{fig:loihi_edge_neurons} (b).

%is why it is thought that sparse data is better suited for spiking neural networks.  

%In Figure \ref{fig:loihi_gray_neurons}, a representative grayscale input image is shown, with the associated neural activities and output predictions plotted as a function of timesteps. The neural activities are shown as spikes as a function of time. 

%In NengoDL, spike probes were placed in each layer on the Loihi chip, and are shown with the first convolutional layer at the top, and the last fully connected layer at the bottom. The probes were set to monitor a sample size of 50 neurons in each layer. %\textcolor{blue}{\st{It can be observed in Figure 6 that with grayscale images, the spiking levels are inconsistent in the input layer. There does not appear to be sufficient spiking activity for the network to make an accurate prediction.}} 
%Because the input data is not sparse, the network has difficulty determining what features are actually important \textcolor{orange}{and thus the overall activity level is much higher, leading to noiser output.} When neurons fire more regularly, it is easier for the network to discern what is actually important, which is why it is thought that sparse data is better suited for spiking neural networks.

%In contrast, focusing on Figure \ref{fig:loihi_gray_neurons}, it can be observed that the first (top) layer has many neurons firing at each timestep. Because of the increase in spiking activity, more spiking information is propagated through the rest of the network. Once the data reaches the last layer, it is able to make the correct prediction on the input data. The sparsity helps the network more effectively determine which features are important and thus improves the overall classification performance of the network. 

\begin{table*}
\caption{Performance results for different edge AI accelerators and Loihi neuromorphic processor.}
\label{table:acc_pdp_all_devices}
\vspace{-2mm}
\centering
\resizebox{0.99\textwidth}{!}{\begin{tabular}{|cl|cccccc|llllllllll|}
\hline
\multicolumn{2}{|c|}{\multirow{2}{*}{Device}} & \multicolumn{6}{c|}{Best Model Architecture} & \multicolumn{2}{c}{\multirow{2}{*}{\begin{tabular}[c]{@{}c@{}}Accuracy (\%)\end{tabular}}} & \multicolumn{2}{c}{\multirow{2}{*}{\begin{tabular}[c]{@{}c@{}}Latency (ms)\end{tabular}}} & \multicolumn{2}{c}{\multirow{2}{*}{FPS}} & \multicolumn{2}{c}{\multirow{2}{*}{\begin{tabular}[c]{@{}c@{}}Power (W)\end{tabular}}} & \multicolumn{2}{c|}{\multirow{2}{*}{\begin{tabular}[c]{@{}c@{}}Energy (mJ)\end{tabular}}} \\ \cline{3-8}
\multicolumn{2}{|c|}{}                        & K1    & K2    & K3    & K4   & FC1   & FC2   & \multicolumn{2}{c}{}                                                                         & \multicolumn{2}{c}{}                                                                        & \multicolumn{2}{c}{}                     & \multicolumn{2}{c}{}                                                                     & \multicolumn{2}{c|}{}                                                                       \\ \hline
\multicolumn{2}{|c|}{Pi}                      & 16    & 24    & -     & -    & 100   & 80    & \multicolumn{2}{c}{96.95}                                                                    & \multicolumn{2}{c}{2.88}                                                                    & \multicolumn{2}{c}{347}                  & \multicolumn{2}{c}{1.56}                                                                 & \multicolumn{2}{c|}{4.49}                                                                   \\
\multicolumn{2}{|c|}{Jetson-L}         & 10    & 28    & -     & -    & 120   & 85    & \multicolumn{2}{c}{97.46}                                                                    & \multicolumn{2}{c}{1.62}                                                                    & \multicolumn{2}{c}{617}                  & \multicolumn{2}{c}{0.91}                                                                 & \multicolumn{2}{c|}{1.47}                                                                   \\
\multicolumn{2}{|c|}{Jetson-H}         & 16    & 24    & -     & -    & 100   & 80    & \multicolumn{2}{c}{95.95}                                                                    & \multicolumn{2}{c}{1.60}                                                                    & \multicolumn{2}{c}{625}                  & \multicolumn{2}{c}{1.34}                                                                 & \multicolumn{2}{c|}{2.14}                                                                   \\
\multicolumn{2}{|c|}{Pi + NCS2}               & 16    & 24    & -     & -    & 100   & 80    & \multicolumn{2}{c}{97.46}                                                                    & \multicolumn{2}{c}{2.35}                                                                    & \multicolumn{2}{c}{425}                  & \multicolumn{2}{c}{2.08}                                                                 & \multicolumn{2}{c|}{4.89}                                                                   \\
\multicolumn{2}{|c|}{Pi + TPU}                & 16    & 32    & -     & -    & 115   & 85    & \multicolumn{2}{c}{97.46}                                                                    & \multicolumn{2}{c}{1.55}                                                                    & \multicolumn{2}{c}{645}                  & \multicolumn{2}{c}{0.77}                                                                 & \multicolumn{2}{c|}{1.19}                                                                   \\
\multicolumn{2}{|c|}{Coral Dev}               & 18    & 24    & -     & -    & 110   & 95    & \multicolumn{2}{c}{96.25}                                                                    & \multicolumn{2}{c}{0.39}                                                                    & \multicolumn{2}{c}{2564}                 & \multicolumn{2}{c}{0.52}                                                                 & \multicolumn{2}{c|}{0.203}                                                                  \\ \hline
\multicolumn{2}{|c|}{Intel Loihi}             & 12    & 22    & 48    & -    & 100   & 85    & \multicolumn{2}{c}{97.40}                                                                    & \multicolumn{2}{c}{35}                                                                      & \multicolumn{2}{c}{28.5}                 & \multicolumn{2}{c}{0.0012}                                                               & \multicolumn{2}{c|}{0.042}                                                                  \\ \hline
\end{tabular}}
\end{table*}

\subsection{latency}
To measure the latency for edge AI accelerators, we perform 40 consecutive inferences on each device and obtain the average inference latency. For the neuromorphic hardware, we measure the propagation delay from the time input spike trains are first applied to the network to the time when output probability is returned. The latency results for all devices are listed in Table \ref{table:acc_pdp_all_devices}. With a 35 millisecond (ms) latency, Loihi is 12$\times$ and 89$\times$ slower than the slowest and fastest edge devices which are Raspberry Pi and Corel Dev board, respectively. Although this increased latency might raise initial concerns, inspecting the achieved frames per second (FPS) rate provides a more nuanced perspective. Notably, all examined devices, including Loihi, fulfill the real-time requirement of 20-30 FPS, underscoring that excessively optimizing latency might not be necessary for many real-world applications such as the facial expression recognition task which is the focus of this work.

%While at first sight it can be concerning, looking at the achieved FPS rate, we can observe that all of the investigated devices including Loihi achieve the real-time requirement of 20 FPS, meaning that the over-optimization of the latency might not be necessary in real-world applications such as facial expression recognition.  

%Because the network is being implemented on neuromorphic hardware, there is a time cost for the physical spikes to propagate through all of the layers of the network. Looking closer at Figure \ref{fig:loihi_edge_neurons}, it takes about \textcolor{blue}{\st{45}} \textcolor{orange}{35} milliseconds for the spiking information to propagate through to the last layer of the network. Because we are concatenating and applying static and unrelated images to the network, there is some delay required for the spiking information to be transmitted through to the output layers. SNNs are better suited for processing real-time information like a video sequence, where each frame of the video is highly correlated with the previous frames \cite{rasmussen2019nengodl}. Additionally, there are several adjustable parameters that can be optimized to improve the response time of the network, on which we have not focused in this paper, leaving it for future developments.

\subsection{Power and Energy}
To measure the power dissipation of edge devices, except for the Jetson Nano, we employed a USB multi-meter, called MakerHawk UM34C, as shown in Fig. \ref{fig:expri_hw}. To attain the dynamic power of the devices, we first measure the idle power of the devices for three minutes without running any of the models, then we run the models for three minutes and measure the total power. By subtracting the idle power from the total power, we can obtain the dynamic power dissipation of the inference operation. For the Jetson Nano, we use its built-in sensors that can measure the CPU and GPU powers separately. The power measurements are provided in Table \ref{table:acc_pdp_all_devices}.

%Due to its inline nature, this measurement device records the whole system's power utilization, including USB I/O. After connecting the power measuring instrument to an input power port, both the idle and running powers are recorded. First, idle power is measured per second for three minutes after connecting the edge devices to the multi-meter without running any models on the devices. Next, we run all models for 3 minutes on each device and measure the dynamic power by subtracting the recorded power from the idle state from the total power dissipated while running the models. To measure the power dissipation of Jetson Nano, we use its three internal sensors, which are located at the board's power input, the GPU, and the CPU. We read the sensors automatically using the \textit{tegrastats} utility. For Jetson Nano, we run each model for three minutes and log the CPU and GPU power consumption.

In addition, %, as provided in Table \ref{tab:loihi_power}, 
we probed the power dissipation of the entire network during the inference operation on the Loihi chip. The total power consumption for each test run was 26.3 mW for the grayscale images and 25.2 mW for the edge-detected images. The dynamic power consumption of the Loihi cores, however, was found to be 2.3 mW for the grayscale images and 1.2 mW for the edge-detected images. The model was able to fit on a single chip, and consumed $91$ cores from the Loihi. This shows that besides the accuracy benefits, edge detection provides roughly 2$\times$ reduction in the dynamic power dissipation. A comparison between Loihi and edge AI accelerators exhibits that Loihi dissipates 1733$\times$ and 433$\times$ less power compared to most power-intensive and least power-intensive devices which are Pi+NCS2 and Coral Dev Board, respectively. Also, from the energy consumption results listed in Table \ref{table:acc_pdp_all_devices}, we can observe that Loihi can achieve 116$\times$ and 4.8$\times$ energy reduction compared to Pi+NCS2 and Coral Dev Board, respectively.

% \begingroup
%     \setlength{\tabcolsep}{6pt} % Default value: 6pt
%     \renewcommand{\arraystretch}{1.5} % Default value: 1
%     \begin{table}
%         \centering
%                 \caption{Power consumption of the Loihi system.}
%         \label{tab:loihi_power}
%         \begin{tabular}{lccc}
%             \textbf{Model}  & \textbf{Grayscale} & \textbf{Edge Detection}\\
%             \cline{1-3}
%             Total Power (W)      & $0.0263$ & $0.0252$ \\
%             Dynamic Power (W) & $0.0023$ & $0.0012$ \\
%             \hline
%         \end{tabular}

%     \end{table}

% \subsection{Quantitative Comparison Between Loihi and Edge AI Accelerators}

% Even though the latency is higher in the SNN due to the propagation time of the spiking signals, the Accuracy/PDP score is significantly higher than other devices at $1745.37$, as shown in Table \ref{table:acc_pdp_all_devices}. The power consumption per inference is much lower on the neuromorphic hardware in comparison to other edge accelerators. This can be attributed to the synaptic response of the neurons, which only fire when necessary. As stated previously, there is a large number of configuration parameters that can be tuned on the Intel Loihi chip, which leaves room for further work in this area. The low power consumption and high classification rate of the Loihi chip show that neuromorphic hardware can achieve state-of-the-art rates of classification while consuming very little power per inference.

\section{Conclusion}
In this paper, we focused on the development and optimization of ML models designed for performing facial expression recognition at the edge. In the first phase of the research, we leveraged hardware-aware neural architecture search to optimize CNN models for deployment on four distinct edge AI accelerators. Among these edge platforms, the coral Dev Board emerged as the best option offering the lowest latency, power dissipation, and energy consumption while maintaining a high level of accuracy. In the second phase, we delved into the conversion of the CNN models into SNNs for deployment on the Loihi neuromorphic processor. To enhance the models' performance within the neuromorphic system, we explored the impact of increasing input image sparsity through an additional pre-processing step involving edge detection. Our findings revealed that edge detection not only reduced power dissipation and energy consumption by decreasing the neural activities within SNN layers but also led to improved accuracy which can be attributed to less noisy inputs. Ultimately, we compared Loihi against edge AI accelerators in terms of accuracy, latency, power, and energy consumption. The comparison results underscored Loihi's remarkable advantages, with approximately two orders of magnitude reduction in power dissipation and one order of magnitude in energy savings compared to edge AI accelerators while maintaining a comparable accuracy. In terms of latency, Loihi lagged behind the edge accelerators. Nonetheless, it managed to fulfill the real-time requirements with a processing speed of more than 28 frames per second.

\bibliographystyle{IEEEtran}

\balance
\bibliography{ref}
%\begin{thebibliography}{1}


%\end{thebibliography}

\end{document}